\documentclass{article}

\usepackage[final]{neurips_2023}

\usepackage[utf8]{inputenc} 
\usepackage[T1]{fontenc}    
\usepackage{hyperref}       
\usepackage{url}            
\usepackage{booktabs}       
\usepackage{amsfonts}       
\usepackage{nicefrac}       
\usepackage{microtype}      
\usepackage{xcolor}         
\usepackage{float,graphicx,multirow,caption}
\title{RAFIC: Retrieval-Augmented Few-shot Image Classification}

\author{%
  Hangfei Lin \qquad Li Miao \qquad Amir Ziai \\
  Stanford University \\
  \tt\small \{hangfei, limiao, amirziai\}@stanford.edu \\
}

\begin{document}
\maketitle

\section{Extended Abstract}
Few-shot image classification is the task of classifying unseen images to one of $N$ mutually exclusive classes, using only a small number of training examples for each class. The limited availability of these examples (denoted as $K$) presents a significant challenge to classification accuracy in some cases.

  To address this, we have developed a method for augmenting the set of $K$ with an addition set of $A$ retrieved images. We call this system Retrieval-Augmented Few-shot Image Classification (RAFIC).
  
  Through a series of experiments, we demonstrate that RAFIC markedly improves performance of few-shot image classification across two challenging datasets. RAFIC consists of two main components: (a) a retrieval component which uses CLIP \cite{schuhmann2022laion}, LAION-5B \cite{schuhmann2022laion}, and faiss \cite{johnson2019billion}, in order to efficiently retrieve images similar to the supplied images, and (b) retrieval meta-learning, which learns to judiciously utilize the retrieved images. Code and data is available at \href{http://github.com/amirziai/rafic}{github.com/amirziai/rafic}.

\subsection{Key Findings}
\begin{enumerate}
    \item \textbf{Using CLIP embeddings leads to vastly superior performance vs. raw pixels.} We show that using CLIP \cite{radford2021learning} embeddings as input features significantly surpasses the performance of raw pixels. For instance, we see accuracy increase from 0.26 to 0.88 in the 10-way rare bird classification task.
    \item \textbf{Zero-shot retrieval using class names is highly effective.} We also found that zero-shot retrieval using class names text embedding is highly effective, achieving over 96\% accuracy in 10-way bird classification, surpassing other methods and showcasing CLIP's familiarity with certain concepts over others.
    \item \textbf{Efficient retrieval-augmentations leads to a boost in accuracy.} The addition of retrieved images significantly improved model accuracy in two challenging few-shot tasks compared to a baseline of logistic regression without meta-training or augmentation. Using Approximate Nearest Neighbors (ANN) search via faiss \cite{johnson2019billion}, we have enabled real-time retrieval and representation extraction across LAION-5B, a repository of 5B+ images.
    \item \textbf{Meta-learning the retrieval strategy further boosts accuracy.} We demonstrate that accuracy can be further improved by using coarse (up or downweighting retrieval as a whole) and fine-grained (up or downweighting individual retrieved images) strategies for meta-learning the retrieval strategy.
    \item \textbf{MAML is more adept at adaptation than ProtoNet.} Training on one task and evaluating on another (cross-evaluation) showed that MAML \cite{chen2021few} adapts well relative to ProtoNet \cite{snell2017prototypical}. Moreover, inclusion of more retrieval images helped with further closing of the performance gap.
\end{enumerate}

Overall, our research validates that retrieval and meta-learning approaches boost performance in few-shot image classification, particularly in cross-task evaluation settings.

\section{Introduction}

\begin{figure}[H]
    \centering
    \setlength{\abovecaptionskip}{0pt}
    \setlength{\belowcaptionskip}{0pt} 
    \includegraphics[width=0.99\textwidth]{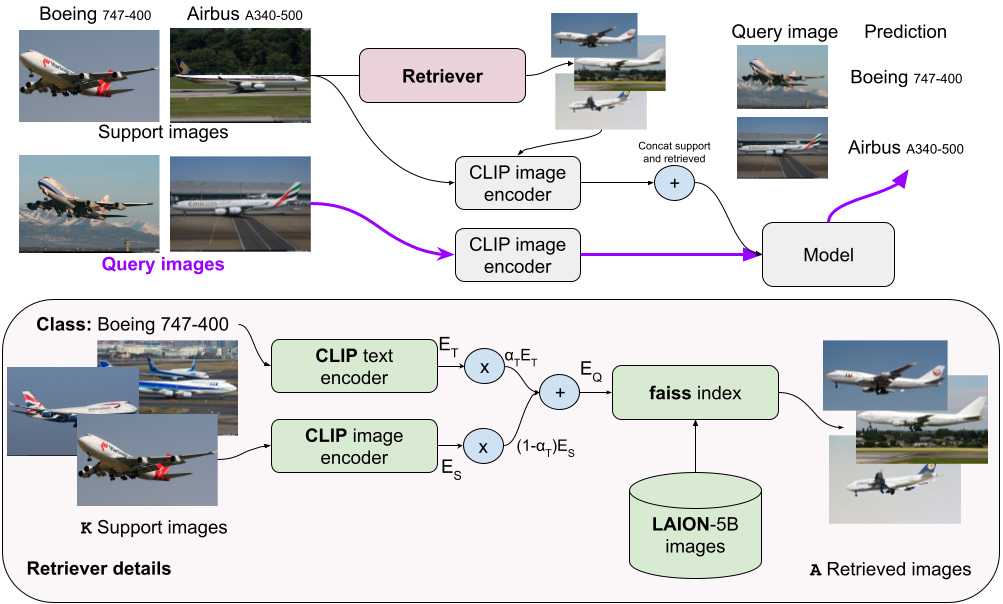}
    \caption{Overview of our proposed system Retrieval-Augmented Few-shot Image Classification (RAFIC). The top diagram shows the high-level view and the bottom one shows the details of the retriever. The top diagram depicts 2-way ($N=2$) classification of aircraft variants. For each class, we provide 1 support image ($K=1$) and ask the model to classify 1 query image ($Q=1$). The retriever uses support images and class labels to retrieve 3 ($A=3$) additional images images. We then concatenate image CLIP embeddings from both support and retrieved images and use that for model training. The model is evaluated on correctly predicting the labels of the query images. Retrieval is done for each class label. We use the CLIP text encoder to embed the class name ($E_T$). We then extract image embeddings for each support image using CLIP, and take the mean ($E_S$). We retrieve the top-$A$ images from LAION-5B using a faiss index and use $E_Q=\alpha_T E_T + (1-\alpha_T)E_S$ as the query, where $0 \le \alpha_T \le 1$. }
    \label{fig:fig1}
\end{figure}

Few-shot image classification\cite{dhillon2019baseline} is a rapidly evolving domain in machine learning, characterized by the challenge of training models to accurately classify new images into a predefined set of categories, despite the availability of only a limited number of samples per class. This approach is especially valuable in fields where acquiring large datasets is impractical or impossible, such as medical imaging \cite{kotia2021few}, drug discovery \cite{vella2022few}, and precision manufacturing \cite{deshpande2020one}. The ability to make accurate predictions from minimal data not only economizes on the resources typically required for training conventional machine learning models but also opens up new possibilities in areas where data is scarce or expensive to acquire.

Our study centers on meta-learning\cite{vilalta2002perspective}, a leading paradigm in solving few-shot image classification. This paradigm focuses on training models that can effectively generalize to new tasks with sparse data. Within this context, we explore various methodologies, including metric-based approaches like ProtoNet\cite{snell2017prototypical}, known for producing feature embeddings through metric learning, and non-metric-based methods such as MAML\cite{finn2017modelagnostic}, which utilizes gradient-based strategies for model parameter optimization. However, these techniques are often limited by the data scarcity challenge inherent in few-shot scenarios.

To mitigate this limitation, data augmentation\cite{ni2021data} has emerged as a popular strategy. This encompasses a range of techniques like image flips, crops, patch erasures, and the introduction of image noise\cite{shorten2019survey}. Furthermore, advanced methods such as generating new images via feature space interpolation\cite{khodadadeh2020unsupervised} or using generative models are being explored\cite{bowles2018gan}. Variations in data augmentation, including augmenting either the support or the query set, have been extensively investigated in the literature \cite{antoniou2017data}.

Our research explores the potential of employing an intelligent retrieval-based data augmentation system. The system searches and fetch additional relevant $A$ examples from external sources, thereby artificially expanding the few-shot dataset from $K$ samples to a more robust size $K + A$. Specifically, our system relies on the large-scale image-text dataset, LAION \cite{schuhmann2022laion}. We use CLIP \cite{radford2021learning} embedding and faiss \cite{johnson2019billion} for image featurization and relevant image retrieval, as illustrated in Figure 1. By retrieving images that are contextually similar to the few-shot examples, we can enrich the training dataset, thereby providing the model with a more comprehensive learning base. We demonstrate that this system can significantly enhance the classification performance of few-shot image classifiers.

Another unique aspect of our approach involves meta-learning the retrieval strategy. This includes adding attention on image similarity, adjusting learning rates on retrieved data, and weighted loss computation. The goal is for the model to discern the most beneficial images for use, a process we believe leads to noticeable improvements. We further explored the generalization capabilities of the models by cross-evaluation the models against a complete different image dataset. We see some gains even in the cross-evaluation scenarios.

The primary contribution of our work lies in introducing retrieval augmentation to few-shot image classification, where we develop an end-to-end system capable of utilizing LAION, faiss and CLIP for data augmentation. Additionally, the meta-learning aspect of the retrieval strategy represents a significant innovation. This methodology's effectiveness across various domains promises potential advancements in applying few-shot image classification in real-world scenarios.

\section{Related Work}

Meta-learning aims to develop models that can quickly adapt to new tasks with limited data and generalize effectively to unseen examples. Within this field, two primary families of approaches have gained prominence: non-metric and metric-based methods. Non-metric methods, exemplified by Model-Agnostic Meta-Learning (MAML)\cite{finn2017modelagnostic}, optimize a model's initial parameters for rapid fine-tuning across diverse tasks. Metric-based approaches, such as  Prototypical Networks\cite{snell2017prototypical}, learn a distance function for classifying instances by comparing them to labeled examples in a support set. Each approach addresses the challenge of adaptation and generalization in unique ways: non-metric methods like MAML optimize for parameter flexibility, while metric-based methods focus on learning effective similarity measures for classification.

While both non-metric and metric-based meta-learning methods have shown promise, they both grapple with the issue of data scarcity. Addressing this challenge has become a focal point in both computer vision and natural language processing (NLP) domains within the realm of meta-learning. A pivotal strategy in this context is data augmentation\cite{ni2021data}, which has been widely recognized for its effectiveness in enhancing meta-learning performance. Our research contributes to this area, focusing on innovative data augmentation techniques to mitigate the limitations posed by limited datasets.

\subsubsection{Meta-learning for image classification}
Various innovative methods have been developed to overcome the challenge of data scarcity in image classification.

\cite{chen2021few} attempt to mitigate the data scarcity problem by retrieving additional images from a base dataset of unlabeled images. Retrieval of similar images is done via training a linear classifier on the support images and then using that classifier to retrieve images with the largest score. This paper presents thorough ablation studies to tease out the impact of several design decisions such as the number of augmented images and model architecture. However, the authors do not experiment with different retrieval strategies. 

\cite{zhang2022tip} leverage both visual and text CLIP \cite{radford2021learning} embeddings to fine-tune a module for improving classification. Similar to this work, we plan to use CLIP as our base visual and textual encoder. However, this work does not use retrieval-augmentation. 

\cite{Zhang_2019_CVPR} present a novel approach to few-shot learning by leveraging saliency-guided hallucination of samples. Their method segments existing images into background and foreground components, which are then reassembled to generate new, synthesized images. This technique is innovative in its utilization of saliency information to enrich the learning dataset. While the method shows promise, it inherently has limitations concerning the diversity of synthesized samples. The performance tends to plateau as the number of unique combinations derived from the background and foreground segments is exhausted. 

One notable approach is identified in \cite{ni2021data} where four distinct modes of augmentation—query, support, task, and shot—are explored. This research also introduces Meta-MaxUp, a method that combines various meta-specific data augmentations, significantly enhancing the performance of prevalent meta-learning algorithms without the need for additional images. 

In supervised image classification, RAC \cite{long2022retrieval} also explore the idea of external augmentation to improve classification accuracy.

Our research builds upon these foundations, introducing retrieval augmentation from large-scale image dataset to image classification in meta-learning. By augmenting the training data and meta-learning the retrieval strategy, we leverage insights from the aforementioned studies. Specifically, RAFIC utilizes the LAION-5B dataset and CLIP embedding, incorporating image and class text embedding for enhanced image retrieval, thus addressing the critical challenges identified in the related work.

\section{Methods}
\label{methods}
This section introduces our proposed method and details the components we have leveraged.

\subsection{Few-shot image classification}
Few-shot image classification is the task of classifying $Q$ query images (per class) belonging to $N$ mutually exclusive classes, given a few examples (i.e. $K$) from each class. It differs from normal image classification in that $K$ is usually very small.

\subsection{Proposed method: Retrieval-Augmented Few-shot Image Classification (RAFIC)}
Fig \ref{fig:fig1} depicts an overview of our proposed system RAFIC. We leverage a retriever to augment the support set with $A$ additional images retrieved from a large scale repository of images. We hypothesize that the inclusion of these additional images, combined with strategies for meta-learning the retrieval strategy, can improve classification accuracy.

\subsection{CLIP for image and text encoding}
OpenAI's CLIP \cite{radford2021learning} was trained on 400M+ image-text pairs. By combining text and image encoders, CLIP learns a shared embedding space where images and text can be directly compared. This makes clip highly efficient and effective at zero-shot retrieval and image/text encoding. RAFIC uses CLIP (ViT-L/14) for encoding images and text.

\subsection{Retriever}
RAFIC's retrieval component uses (1) CLIP, (2) a large corpus of 5B images (LAION-5B), and (3) an efficient retrieval index (faiss), for retrieving and augmenting support images.

\subsubsection{LAION-5B}
LAION-5B \cite{schuhmann2022laion} is a corpus of 5B+ image-text pairs. At the time of this writing, this dataset is the largest open-source repository of images. RAFIC retrieves images from LAION-5B that are most similar to the supplied support images.

\subsubsection{Retrieval index (faiss)}
RAFIC retrieves similar images for each class. Brute-force retrieval search across 5B+ images is prohibitively expensive. Therefore, we resort to Approximate Nearest Neighbor (ANN) search via faiss \cite{johnson2019billion}. We use cosine similarity as our retrieval score.

\subsubsection{Retrieval mechanics}
Given a set of $K$ support images for class name $T$ (a string), we first use the CLIP text encoder to encode the string formatted as "photo of \{T\} \{dataset\}", we call this $E_T$. An example of this string is "photo of a Boeing 747-400 aircraft", where $T$ is "Boeing 747-400" and dataset is "aircraft".

Next, we encode each image using the CLIP image encoder and take the mean of the image embeddings to get $E_S$. Then, we compute $E_Q = \alpha_T E_T + (1-\alpha_T)E_S$ (where $0 \le \alpha_T \le 1$) and use $E_Q$ to search against the faiss index.

\subsection{Model}
Image embeddings for the support and retrieved images are concatenated into a combined matrix $X$. Then, $X$ is passed into a learning model. This model is ultimately responsible for predicting query image labels. As we will see shortly, the model may or may not use $X$. We considered a few models that we will now describe.

\subsubsection{Logistic Regression (LR)}
We learn a Logistic Regression classifier using $X$ for each task. In other words, a new classifier is trained for each $N$-way few-shot classification task, and there's no opportunity for meta-learning across tasks. We use the scikit-learn \cite{pedregosa2011scikit} implementation of LR.

\subsubsection{Model-Agnostic Meta-Learning (MAML)}
MAML \cite{finn2017modelagnostic} is a widely used meta-learning technique. MAML involves an inner-loop, which can adapt to the task, and an outer loop, which updates the model parameters while training. We use an an MLP on top of CLIP embeddings with ReLU activation.

\subsubsection{Prototypical Networks (ProtoNet)}
ProtoNet \cite{snell2017prototypical} learns by creating "prototypes", which is an average of the support images per task. To classify unseen images, ProtoNet simply compares their features to these prototypes, and assigns them to the closest matching class. Similar to MAML, we use an MLP with ReLU on top of CLIP embeddings.

\subsubsection{Zero-shot (ZS) classification using class names}
This model is special in that it does not use the support images and therefore does not use retrieval (i.e. ignores $X$). We follow this approach:
\begin{enumerate}
\item Extract CLIP embeddings for query images and the class names.
\item For each query image, calculate the cosine similarity between image embeddings and each of the class text embeddings.
\item Output the class name with the highest score.
\end{enumerate}

\subsection{Retrieval meta-learning}
In our retrieval augmentation, we employ two following two categories of meta-learning approaches:
\begin{itemize}
\item \textbf{Coarse-grained.} We use two techniques: (1) meta-learn separate learning rates for support and retrieval component, allowing the learning process to over or under-emphasize retrieval as a whole, and (2) weighted loss computation, we use similarity score to overemphasize retrieved images with higher visual alignment in the learning process. We have only implemented this method for MAML.
\item \textbf{Fine-grained.} For each retrieved image, we append the retrieval cosine similarity score to the image CLIP embeddings. For support and query images, we append the value of 1.
\end{itemize}

\section{Experiments}
In this section, we conduct a series of experiments to study our proposed system.

\subsection{Datasets}
We use two datasets for our experiments.

\subsubsection{Birds}
The Caltech-UCSD Birds dataset \cite{wah2011caltech} has a total of 11,788 images from across 200 classes of rare birds. The dataset is split into 70\%, 15\%, and 15\% for train, validation, and test respectively.

\subsubsection{Aircraft}
The FGVC-Aircraft dataset \cite{maji2013fine} contains 10,200 images of aircraft across 102 different aircraft model variants. Each variant belongs to one of 29 manufacturers. For instance, Boeing has 22 different variants such as 747-100 and 777-200. To avoid leakage across splits, we partitioned the data by manufacturer, which yielded approximately 65\%, 20\%, and 15\% for train, validation, and test.

\subsection{Experimental setup}
For all experiments, we picked the following values:
\begin{itemize}
    \item \textbf{$N=10$}, which implies random guessing accuracy of 0.1.
    \item \textbf{$K=1$}, accuracy tends to improve with more support images. Picking the smallest possible value left more room for observing the impact of our proposed method.
    \item \textbf{$Q=5$}, our early experiments indicated that the results are not very sensitive to this value.
    \item \textbf{$\alpha_T=0.8$}, in our early experiments, we observed that putting more weight on the class name text embedding $E_T$ lead to better results relative to the mean image embeddings $E_S$.
\end{itemize}

We ran experiments using 5 different seeds and observed the variance in the validation set. In most cases the standard deviation is less than 1 percentage point. We report accuracy on the test set.

For MAML and ProtoNet, we train with batch size 8 and up to a maximum of 200 training steps (both methods converge earlier).  For ProtoNet we used a single-layer linear layer with 64 hidden dimensions, but also tried 2-layer MLPs and using 32, 128, 256 neurons, and observed very little impact. For MAML, we use inner and outer learning rate of 0.04 and 0.001 respectively, used 100 inner steps, and a 2-layer MLP with 128 and 32 hidden neurons.

\subsubsection{Compact index construction for efficient experimentation}
To enable efficient experimentation, we opted to precompute a compact index. This approach eliminated the need for real-time embedding extraction. We constructed this index by (1) retrieving the top-20 images from LAION-5B for each image in the dataset using the image embedding, (2) retrieving the top-100 images for each class label, (3) taking the union of all images in the previous step, and (4) extracting these embeddings and constructing a new compact faiss index for retrieval.

Using this approach, we are guaranteed to only retrieve precomputed images. We ran a few experiments and validated that recall was very close to 1 in the majority of cases relative to using the full index.

\subsection{Experiment 0: CLIP vs. raw pixels}
We evaluated the effectiveness of using CLIP embeddings vs. raw pixels as input features for few-shot image classification using a basic MAML setup. CLIP embeddings achieved a remarkable accuracy of 0.88 on the birds dataset, far surpassing the performance of raw image bytes, which only reached 0.26. This clearly demonstrates the significant advantage of using CLIP embeddings, which capture richer semantic information and lead to significantly improved classification performance. To ensure consistent and optimal performance across all experiments, we use CLIP embedding as the default representation for images.

\subsection{Experiment 1: augmenting support images via retrieval}

\begin{figure}[H]
    \centering
    \setlength{\abovecaptionskip}{0pt}
    \setlength{\belowcaptionskip}{0pt} 
    \includegraphics[width=0.99\textwidth]{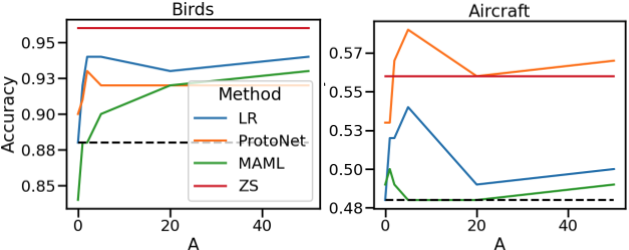}
    \caption{Accuracy as a function of number of retrieved images $A$. We use $A=\{0,1,2,5,20,50\}$. Baseline is the dashed black line, which is LR with $A=0$. We see that accuracy improves as we increase $A$. The red line (ZS) does not use support or retrieved images and only uses class label name embeddings. We see that no method can beat ZS for birds, but ProtoNet outperforms it for aircraft.  }
    \label{fig:exp1}
\end{figure}

This experiment focuses on the effect of adding retrieved images. Fig \ref{fig:exp1} shows accuracy as a function of increasing the number of retrieved images (starting at 0).

We observe that retrieval boosts performance. However, performance plateaus as $A$ increases in most cases. Both MAML and ProtoNet beat the LR baseline (i.e. LR at $A=0$), but ZS beats all methods for birds without referencing support images or additional retrieved images.

This experiment suggests that CLIP has seen more examples of rare birds with rich textual descriptions from across the internet vs. aircraft. Also, aircraft variants are harder to visually distinguish and retrieval and meta-learning appear to boost accuracy more as a consequence.

\subsection{Experiment 2: meta-learning the retrieval strategy}
We start with the observation that retrieved images are not all equally relevant. Ideally, the retrieval strategy should learn to put the correct emphasis on retrieval, and not blindly treat all images equally.

As described in the \nameref{methods} section, we consider techniques to address fine and coarse-grain retrieval meta-learning. We tabulate the effect of enabling both of these vs. no retrieval meta-learning in Table \ref{table:exp2} (results for $A=50$ are flat, which we exclude to save space).

\begin{table}[H]
\centering
\begin{tabular}{|l|l|l|l|l|l|}
\hline
\multicolumn{1}{|c|}{\textbf{Dataset}} & \multicolumn{1}{c|}{\textbf{Method}} & \multicolumn{1}{c|}{\textbf{A=1}} & \multicolumn{1}{c|}{\textbf{A=2}} & \multicolumn{1}{c|}{\textbf{A=5}} & \multicolumn{1}{c|}{\textbf{A=20}} \\ \hline
\multirow{2}{*}{Birds}                 & MAML                                 & 0.88 / 0.88                       & 0.88 / \textbf{0.89}                       & 0.90 / \textbf{0.91}                       & 0.92 / 0.92                        \\ \cline{2-6} 
                                       & ProtoNet                             & 0.91 / \textbf{0.94}                       & 0.93 / 0.93                       & 0.92 / \textbf{0.93}                       & 0.92 / \textbf{0.93}                        \\ \hline
\multirow{2}{*}{Aircraft}              & MAML                                 & 0.50 / \textbf{0.53}                       & 0.49 / \textbf{0.53}                       & 0.48 / \textbf{0.52}                       & 0.48 / \textbf{0.51}                        \\ \cline{2-6} 
                                       & ProtoNet                             & 0.53 / \textbf{0.60}                       & 0.57 / \textbf{0.63}                       & 0.59 / \textbf{0.61}                       & 0.56 / 0.56                        \\ \hline
\end{tabular}
\caption{Effect of meta-learning the retrieval strategy. Results are in x / y format, where y is accuracy with and x is without meta-learning. Cells with an improvement are bolded, and we can see an improvement in most cases, especially for the aircraft dataset. }
\label{table:exp2}
\end{table}

Improvements are larger for smaller values of $A$, ProtoNet enjoys a larger boost, and aircraft benefits more than birds (consistent with experiment 1). Note that ProtoNet only uses fine-grained (i.e. appending cosine similarity of retrieved images to CLIP embeddings). MAML uses both fine and coarse-grained. Ablating fine-grain from MAML led to losing almost all of the benefits. Using only fine-grained for aircraft actually led to an improvement for $A \ge 2$, namely 0.55, 0.53, and 0.53 for $A$ equal to 2, 5, and 20 respectively.

\subsection{Experiment 3: cross-evaluation}
In this experiment, we meta-train methods on one dataset and evaluate it on the other dataset. Obviously, we expect methods to performance worse when not trained on the evaluation task. This experiment is intended to answer two questions:

\begin{enumerate}
    \item How much performance drop do we see? Another way of asking this question is: can one method recover better from not having been trained on the evaluation task?
    \item Do we see a gain with meta-learning the retrieval strategy (i.e. what we tested in experiment 2)?
\end{enumerate}

\begin{figure}[H]
    \centering
    \setlength{\abovecaptionskip}{0pt}
    \setlength{\belowcaptionskip}{0pt} 
    \includegraphics[width=0.99\textwidth]{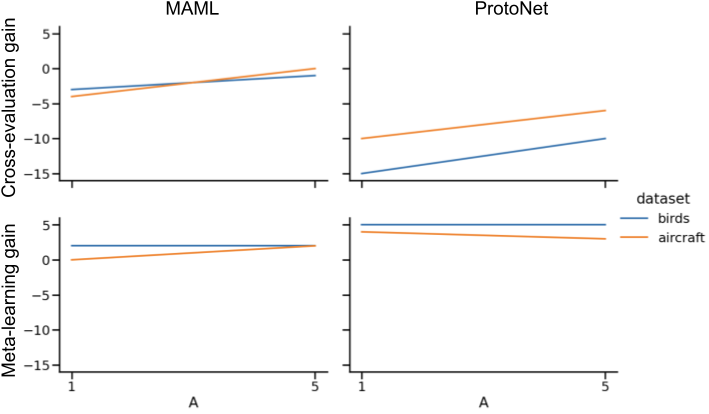}
    \caption{Cross-evaluation results for experiment 3. The y-axis for all four graphs is the absolute difference in accuracy times 100. For example, if accuracy with and without a treatment is 0.82 and 0.80 respectively, the difference (i.e. gain) is 2 percentage points. The first row focuses on question 1 and measures the drop (i.e. negative gain) in accuracy when we cross-evaluate vs. training and evaluating on the same dataset. The second row focuses on question two, and shows the accuracy gain of using retrieval meta-learning (described in experiment 2) vs. not using it (both are cross-evaluated). We omitted results for $A>5$ as they ended up being very similar to $A=5$, for the sake of brevity. }
    \label{fig:exp3}
\end{figure}

Fig \ref{fig:exp3} depicts the results of our experiments for answering these two questions.

We notice that MAML can better adapt much vs. ProtoNet. This finding makes sense in light of understanding how MAML works, and especially given the fact that we use 100 inner steps for adaptation. ProtoNet, on the other hand, has no opportunity to adapt since the learned parameters are fixed at evaluation. Also, we see that there's less of a drop as we go from 1 to 5. Additional retrieved images appear to make tasks more transferable across both methods.

Focusing on the second row in Fig \ref{fig:exp3}, we see that meta-learning the retrieval strategy plays a positive role, with the effect being larger for ProtoNet.

This experiment introduces an additional angle to our study. Although we saw MAML consistently underperform the simpler ProtoNet in previous experiments, we can see that it's more adaptable and might be a better choice for meta-learning across radically different tasks. We should note that MAML involves more hyperparameters relative to ProtoNet, and we might've been able to see similar or better performance to ProtoNet with thorough hyperparameter tuning.

\section{Conclusion and future work}
Our study suggests that retrieval-augmentation using a large-scale image repository (LAION-5B), efficient image-text encoders (CLIP), and efficient retrieval indices (faiss), can benefit few-shot image classification. We have also demonstrated that meta-learning is beneficial for faster adaptation and higher performance when labeled data is hard or expensive to collect.

For future work, we aim to broaden our research by incorporating a more diverse range of datasets from different domains, in order to better assess the validity of our method. Additionally, we have only scratched the surface when it comes to meta-learning the retrieval strategy. For instance, having observed the benefits from including textual descriptions, we plan to devise and experiment with methods that further leverage natural language.

\section{Contributions}
\begin{enumerate}
    \item \textbf{Hangfei} conducted data exploration, optimized CLIP embeddings extraction, and build the data loader for retrieval, training and evaluation.
    \item \textbf{Li} developed the training and evaluation framework. He also designed and implemented meta-learning methods for retrieval.
    \item \textbf{Amir} built and optimized the faiss index for retrieval and designed and conducted the experiments.
\end{enumerate}

We deviated from our original plan as we realized that efficient retrieval is key to fast experimentation, and that developing and optimizing the retrieval index requires additional effort. This led to Li taking over the development of new meta-learning methods instead of Amir.

\section{Acknowledgements}
We would like to thank Yoonho Lee, whose insightful discussions and guidance have been instrumental in shaping and refining this work. Additionally, our thanks go to Chelsea Finn for her valuable advice in the selection of datasets and providing early feedback in this research direction.


\medskip

\bibliographystyle{plainnat}
\bibliography{references}
\end{document}